\documentclass[sigconf]{acmart}

\usepackage{amsmath}
\usepackage{algorithmic}
\usepackage{graphicx}
\usepackage{textcomp}
\usepackage{xcolor}
\usepackage{colortbl}
\usepackage{enumerate}
\usepackage{multirow}
\usepackage{hyperref}
\usepackage{dirtytalk}

\AtBeginDocument{%
  }

\setcopyright{None}

\renewcommand\footnotetextcopyrightpermission[1]{} 
    
\begin{document}

\title{Harnessing PU Learning for Enhanced Cloud-based DDoS Detection: A Comparative Analysis}

\author{Robert Dilworth}
\affiliation{
  \institution{Mississippi State University}
  \city{Starkville}
  \state{Mississippi}
  \country{USA}
}
\email{rkd103@msstate.edu}

\author{Charan Gudla}
\affiliation{
  \institution{Mississippi State University}
  \city{Starkville}
  \state{Mississippi}
  \country{USA}
}
\email{gudla@cse.msstate.edu}

\begin{abstract}
    This paper explores the application of Positive-Unlabeled (PU) learning for enhanced Distributed Denial-of-Service (DDoS) detection in cloud environments. Utilizing the \texttt{BCCC-cPacket-Cloud-DDoS-2024} dataset, we implement PU Learning with four machine learning algorithms: XGBoost, Random Forest, Support Vector Machine, and Na\"{i}ve Bayes. Our results demonstrate the superior performance of ensemble methods, with XGBoost and Random Forest achieving \(F_{1}\) Scores exceeding 98\%. We quantify the efficacy of each approach using metrics including \(F_{1}\) Score, ROC AUC, Recall, and Precision. This study bridges the gap between PU Learning and cloud-based anomaly detection, providing a foundation for addressing Context-Aware DDoS Detection in multi-cloud environments. Our findings highlight the potential of PU Learning in scenarios with limited labeled data, offering valuable insights for developing more robust and adaptive cloud security mechanisms.
\end{abstract}



\keywords{
    Cloud Computing and Security, 
    Cybersecurity, 
    Distributed Denial-of-Service (DDoS), 
    Machine Learning, 
    Positive-unlabeled (PU) Learning,
    Negative-Unlabeled (NU) Learning,
    Anomaly Detection, 
    XGBoost, 
    Random Forest, 
    Support Vector Machine, 
    Na\"{i}ve Bayes
}

\maketitle

\pagestyle{plain}

\section{Introduction}

    The rapid evolution of cloud computing has brought about unprecedented challenges in cybersecurity, particularly in the realm of Distributed Denial-of-Service (DDoS) attacks. As these attacks grow in sophistication and frequency, traditional detection methods often fall short. This paper explores the potential of Positive-Unlabeled (PU) learning, a semi-supervised machine learning technique, in revolutionizing DDoS detection within cloud environments.
    
    PU Learning offers a unique approach to anomaly detection by leveraging a set of known positive instances and a pool of unlabeled data. This method is particularly suited to scenarios where obtaining comprehensive labeled datasets is challenging or impractical. In the context of cloud security, PU Learning could potentially overcome limitations associated with traditional supervised learning approaches, which often struggle with the dynamic and evolving nature of DDoS attacks.

\subsection{Research Objectives}

    This study aims to:
    
    \begin{enumerate}
        \item Investigate the efficacy of PU Learning in identifying anomalous cloud traffic, with a specific focus on DDoS attacks.
        \item Evaluate the impact of various underlying machine learning methods on PU Learning performance in the context of DDoS detection.
        \item Contribute to the nascent field of Context-Aware DDoS Detection by exploring PU Learning's adaptability to multi-cloud environments.
    \end{enumerate}

\subsection{Paper Structure}

    The remainder of this paper is organized as follows:
    
    \begin{itemize}
        \item Section 2 provides a comprehensive overview of PU Learning, its mechanisms, and potential applications.
        \item Section 3 explores the theoretical applications of PU Learning in cloud computing environments.
        \item Section 4 delves into anomaly-based DDoS detection strategies.
        \item Section 5 details our experimental design, including dataset selection and implementation of various PU Learning approaches.
        \item Section 6 presents and discusses our findings on the effectiveness of PU Learning in DDoS detection.
        \item Section 7 compares our PU Learning approach with a state-of-the-art DDoS detection model.
        \item Section 8 reframes the task of DDoS detection as an NU Learning problem and analyzes the performance of the original PU configuration against the reconceptualized NU setup.
        \item Section 9 concludes the paper and suggests directions for future research.
    \end{itemize}

\subsection{Paper Contributions}

    This work unites two previously unrelated concepts: the machine learning approach of PU Learning and cloud anomaly detection with an emphasis on Distributed Denial-of-Service (DDoS) attacks. Based on the lack of published literature, there is little research that has adequately addressed the utility of PU Learning in a cloud computing environment. Furthermore, this work will:

    \begin{itemize}
        \item Quantify the predictive and detective efficacy of PU Learning implementations using common machine learning methods: Na\"{i}ve Bayes (NB), Support Vector Machine (SVM), Random Forest (RF), and XGBoost.
        \item Provide the necessary groundwork to address Context-Aware DDoS Detection, a problem we have identified in multi-cloud environments.
        \item Demonstrate how PU Learning may play a crucial role in distinguishing between legitimate traffic spikes and cyber attacks, by leveraging its ability to extrapolate insights from datasets with limited or poor labeling.
    \end{itemize}

\section{Understanding PU Learning}

    PU Learning, or Positive-Unlabeled learning, is a semi-supervised machine learning approach that addresses scenarios where only positive and unlabeled data are available for training. This section delves into the fundamentals of PU Learning and its potential applications across various domains.

    \subsection{Core Principles of PU Learning}
        
        In PU Learning, the training data consists of two components:
        
        \begin{enumerate}
            \item Positive (P) examples: Instances known to belong to the positive class.
            \item Unlabeled (U) examples: Instances with unknown class labels (positive or negative).
        \end{enumerate}
        
        The primary objective of PU Learning is to construct a classifier capable of accurately predicting class labels for unlabeled examples, given only positive examples and unlabeled data. This approach differs significantly from traditional supervised learning methods, which require both positive and negative labeled examples during the training process.
    
    \subsection{PU Learning Techniques}
        
        PU Learning methods can be broadly categorized into three main approaches \cite{Bekker2020}:
        
        \begin{enumerate}
            \item Two-step techniques
            \item Biased learning
            \item Class prior incorporation
        \end{enumerate}
        
        \subsubsection{Two-step Techniques}
        
            This approach involves:
            
            \begin{enumerate}[a)]
                \item Identifying reliable negative examples
                \item Learning based on labeled positives and reliable negatives
            \end{enumerate}
        
        \subsubsection{Biased Learning}
        
            This method treats PU data as fully labeled data with class label noise for the negative class.
        
        \subsubsection{Class Prior Incorporation}
        
            This technique modifies standard learning methods by applying mathematics from the Selected Completely At Random (SCAR) assumption--a sampling method where individuals or items are selected randomly from a population without any specific criteria or bias--utilizing the provided class prior.
            
    \subsection{Applications of PU Learning}
        
        PU Learning has shown promise in various fields, including:
        
        \begin{enumerate}
            \item Information retrieval \cite{Kumagai2024}
            \item Bioinformatics \cite{Zhang2024}
            \item Computer vision \cite{Long2024}
            \item Natural language processing \cite{Duan2024}
        \end{enumerate}
        
        Its ability to handle scenarios with limited labeled data makes it particularly valuable in domains where obtaining comprehensive labeled datasets is challenging or resource-intensive.

    \subsection{Assumptions and Limitations}

        According to \textit{Bekker et al.} \cite{Bekker2020}, the common assumptions about data distributions in PU Learning are:
        
        \begin{itemize}
            \item All unlabeled examples are negative
            \item The classes are separable
            \item The classes have a smooth distribution
        \end{itemize}
        
        However, these assumptions may not always hold in real-world scenarios, particularly in the context of cloud computing and DDoS detection. For instance:
        
        \begin{itemize}
            \item The assumption that all unlabeled examples are negative may lead to a \say{close-world assumption,} which, while simplistic, can be serviceable in certain contexts. This means that in situations where it is difficult or impractical to obtain labels for all examples, treating unlabeled data as negative can simplify the analysis and decision-making process, even though it may not accurately reflect the true nature of the data.
            \item The separability of classes is often clear in log data and traffic patterns, where deviations from normal behavior are typically distinct.
            \item The assumption of smooth distribution may not always apply, as log data trends can be irregular.
        \end{itemize}

\section{PU Learning in Cloud Computing and Security}

    The unique characteristics of PU Learning make it a promising candidate for addressing various challenges in cloud computing. This section explores potential applications of PU Learning within these domains, with a particular focus on anomaly detection and DDoS mitigation.

    \subsection{Anomaly Detection in Cloud Infrastructure}
    
        PU Learning can be employed to identify unusual behavior patterns in cloud environments, such as:
        
        \begin{enumerate}
            \item Suspicious network traffic
            \item Unauthorized access attempts
            \item Atypical resource utilization patterns
        \end{enumerate}
        
        By training a PU Learning model on known \say{positive} (malicious or anomalous) examples and \say{unlabeled} (potentially benign or normal) data, the system can learn to flag potential security threats or system malfunctions effectively.
    
    \subsection{Vulnerability Detection}
    
        PU Learning can aid in proactively identifying vulnerabilities in cloud-based applications, services, or infrastructure. The model can be trained on known \say{positive} examples of vulnerabilities and \say{unlabeled} data, enabling security teams to address potential weaknesses before they can be exploited.
    
    \subsection{Malware Detection}
    
        In cloud environments, PU Learning can be utilized to detect malware by training models on known malware samples (positive examples) and unlabeled data that may or may not contain malware. This approach can enhance the protection of cloud-based systems and data against evolving malware threats.
    
    \subsection{User Behavior Analysis}
    
        PU Learning can be applied to analyze user behavior patterns in cloud environments, including:
        
        \begin{enumerate}
            \item Login patterns
            \item Resource usage
            \item Access trends
        \end{enumerate}
        
        This analysis can help identify anomalous user behavior that may indicate account compromise, insider threats, or other security risks.
    
    \subsection{Cloud Resource Optimization}
    
        While not directly related to security, PU Learning can contribute to optimizing cloud resource allocation and utilization. By training models on known examples of efficient resource usage and unlabeled data, the system can learn to identify and recommend ways to optimize cloud resource utilization, potentially reducing costs and improving overall cloud performance.

\section{Anomaly-based DDoS Detection: A Focus on PU Learning}

    This section explores the application of PU Learning in the context of anomaly-based DDoS detection, comparing it with traditional statistical and machine learning approaches \cite{Osanaiye2016}.

    \subsection{Categories of Anomalies}
    
        Anomalies in network traffic can be classified into three main categories:
        
        \begin{enumerate}
            \item \textit{Point anomalies}: Individual data instances considered anomalous compared to the rest of the data (e.g., application-bug level attacks resulting in DoS).
            \item \textit{Contextual anomalies}: Data instances anomalous in specific contexts but not in others, determined by the dataset structure.
            \item \textit{Collective anomalies}: Groups of data instances flagged as anomalous with respect to the entire dataset (e.g., DDoS flooding attacks).
        \end{enumerate}

    \subsection{Traditional Approaches to Anomaly Detection}
    
        \subsubsection{Statistical Anomaly Detection}
        
        This method involves:
        
        \begin{enumerate}[a)]
            \item Compiling statistical features of normal traffic to generate a baseline pattern
            \item Comparing incoming traffic with the baseline using statistical inference tests
            \item Determining the legitimacy of behavior without prior knowledge of normal system activities
        \end{enumerate}
    
        \subsubsection{Machine Learning Anomaly Detection}
        
        This approach utilizes:
        
        \begin{enumerate}[a)]
            \item Training algorithms on datasets of normal behavior to learn patterns and characteristics
            \item Identifying instances that deviate significantly from learned normal behavior
            \item Common techniques include one-class support vector machines, isolation forests, autoencoders, clustering, and density estimation
        \end{enumerate}

    \subsection{PU Learning for DDoS Detection}
    
        PU Learning offers a unique approach to DDoS detection by leveraging:
        
        \begin{enumerate}
            \item Known positive examples of DDoS traffic
            \item Unlabeled traffic data that may contain both normal and malicious patterns
        \end{enumerate}
        
        This method can potentially overcome limitations of traditional approaches, particularly in scenarios where comprehensive labeled datasets are unavailable or impractical to obtain.

    \subsection{Statistical and Machine Learning Anomaly Detection Methods}

        Based on \textit{Shafi et al.'s} \cite{Shafi2024} prior work, we scrutinize and employ four well-established machine learning algorithms: Na\"{i}ve Bayes (NB), Support Vector Machine (SVM), Random Forest (RF), and XGBoost.
        
        \begin{itemize}
            \item \textit{XGBoost}: A supervised gradient boosting decision tree model, falling under the machine learning anomaly detection category.
            \item \textit{Na\"{i}ve Bayes (NB)}: A statistical approach using probability and Bayesian inference, categorized under statistical anomaly detection.
            \item \textit{Support Vector Machines (SVMs)}: A machine learning technique, particularly the one-class variant used for unsupervised anomaly detection.
            \item \textit{Random Forests (RF)}: An ensemble of decision tree models, classified under machine learning anomaly detection.
        \end{itemize}
        
        It is important to note that PU Learning is distinct from these supervised learning algorithms. Specifically, PU Learning is not a technique like XGBoost, Support Vector Machines (SVMs), Random Forests, or Bayesian methods. \textit{PU Learning is a specific type of semi-supervised learning problem.} While XGBoost, Random Forests, and SVMs require both positive and negative examples during training, PU Learning leverages only positive and unlabeled data. This unique characteristic makes PU Learning particularly suitable for scenarios where obtaining negative examples is difficult or expensive, such as in the case of DDoS detection in cloud environments.

\section{Experimental Design and Methodology}

    This section outlines our approach to evaluating the effectiveness of PU Learning in DDoS detection within cloud environments.
    
    \subsection{Dataset Selection}
        We utilize the \texttt{BCCC-cPacket-Cloud-DDoS-2024} dataset \cite{Kaggle2024}, developed by \textit{Shafi et al.} \cite{Shafi2024}, which addresses 15 identified weaknesses in existing DDoS datasets. This comprehensive dataset includes:
        
        \begin{enumerate}
            \item Over eight benign user activities
            \item 17 DDoS attack scenarios
            \item More than 300 extracted features from network and transport layers
        \end{enumerate}

    \subsection*{Dataset Preprocessing}
    
        To prepare the dataset for machine learning algorithms, we follow a structured preprocessing approach. The primary goals of this preprocessing include handling missing values, converting categorical variables into numerical representations, managing extreme values, and ensuring that all data types are compatible for model input.
        
        \begin{enumerate}
            \item \textit{Loading the Dataset}: The first step involves loading the dataset from a text file. We utilize the \texttt{pandas} library to read the CSV-formatted data, ensuring that no header row is considered, which allows us to treat all rows uniformly. Before evaluating the DDoS dataset, we first had to adjust the \texttt{.csv} file. Specifically, we created a new column named \(y\) and populated it with either a 1 or 0 based on the value stored in the \texttt{label} column. The \texttt{label} column identified the row of collected data as either \texttt{Benign} or \texttt{!Benign}; we populated the new column with a value of 1 for \texttt{Benign} and 0 for \texttt{!Benign}. Likewise, we stripped the column names from the \texttt{.csv} and converted it into a text file.
        
            \item \textit{Handling Missing Values}: After loading the data, we address any missing values within the dataset. Empty entries, denoted by consecutive commas or empty strings, are replaced with \texttt{NaN} (Not a Number) values. This replacement facilitates easier handling of missing data in subsequent steps, as \texttt{NaN} is a standard placeholder in \texttt{pandas} for missing or undefined values.
        
            \item \textit{Identifying and Processing Categorical Variables}: To convert these categorical variables into a format suitable for machine learning models, we employ a hashing function. This function takes each categorical value, converts it into a string, encodes it, and generates a fixed-size numerical representation. By applying this hashing function, we ensure that each categorical variable is transformed into a unique integer, thus eliminating any string-based entries from the dataset.
        
            \item \textit{Managing Extreme Values}: As part of the preprocessing, we also need to address any extreme values in the dataset that may skew results or lead to errors during model training. We define a threshold for what constitutes an \say{extreme} value. For any numeric entry that exceeds this threshold, we replace the value with a hashed representation of the original value. This transformation helps maintain numerical stability while ensuring that extreme values do not negatively impact the learning algorithms.
        
            \item \textit{Converting to Numeric Types}: With categorical variables hashed and extreme values managed, we convert all remaining data in the \texttt{DataFrame} to numeric types. This conversion is done using the \texttt{pandas} function that forces any non-convertible values to be set as \texttt{NaN}. After this conversion, we fill any remaining \texttt{NaN} values with a specified default value, such as \texttt{0}.
        
            \item \textit{Final Data Preparation}: Finally, we ensure that all entries in the \texttt{DataFrame} are of integer type, making the dataset ready for input into machine learning models. This final conversion confirms that no string or non-numeric data remains, and that the dataset is in a uniform format suitable for model training and evaluation.
        \end{enumerate}
        
        Through these preprocessing steps, we transform the raw dataset into a clean, structured format that is suitable for use with machine learning algorithms, enhancing both the accuracy and efficiency of subsequent analyses. The adjusted dataset, as a result of the above preprocessing, consists of \textit{700,775 rows} and \textit{325 columns} (or \textit{227,751,875 cells}).
    
    \subsection{PU Learning Implementation}
    
        We adapt the PU Learning implementation \cite{GitHub2020} by \textit{Agmon} \cite{Agmon2020}, based on the work of \textit{Bekker et al.} \cite{Bekker2020}. The process involves:
        
        \begin{enumerate}
            \item Training a classifier to predict the probability of a sample being labeled
            \item Using the classifier to predict the probability of positive samples being labeled
            \item Predicting the probability of a given sample being labeled
            \item Estimating the probability of a sample being positive
        \end{enumerate}

    \subsection*{Code Overview}
    
        The table, \autoref{tab:code_overview}, provides a summary of the key functions and classes within our code, which was designed to evaluate multiple machine learning models on the \texttt{BCCC-cPacket-Cloud-DDoS-2024} dataset \cite{Kaggle2024} using a Positive-Unlabeled (PU) learning approach.
        
        \begin{table*}[h!]
            \centering
            \begin{tabular}{|p{8cm}|p{8cm}|}
                \hline
                \rowcolor{black} \textcolor{white}{\textbf{Function/Class}} & \textcolor{white}{\textbf{Description}} \\
                \hline
                \rowcolor{lightgray} \texttt{ModelEvaluator} & Class that handles the evaluation of machine learning models based on metrics like \(F_{1}\), ROC, Recall, and Precision. \\
                
                \texttt{evaluate\_results(y\_test, y\_predict)} & Static method that evaluates model performance and returns a dictionary with \(F_{1}\) Score, ROC AUC score, Recall, and Precision. \\
                
                \rowcolor{lightgray} \texttt{PULearning} & Class implementing the Positive-Unlabeled learning approach, allowing different classifiers to be used. \\
                
                \texttt{fit\_PU\_estimator(X, y, hold\_out\_ratio)} & Fits the model using PU Learning by holding out a portion of the positive samples for estimation. \\
                
                \rowcolor{lightgray} \texttt{predict\_PU\_prob(X, prob\_s1y1)} & Predicts probabilities using the fitted PU estimator, adjusting for the estimated probability of positive samples. \\
                
                \texttt{compare\_models(models, x\_train, y\_train, x\_test, y\_test)} & Compares multiple machine learning models, fitting each to the same dataset and evaluating performance metrics. \\

                \rowcolor{lightgray} \texttt{evaluate\_pu\_model(model\_class, x\_train, y\_train, x\_test, y\_test, hold\_out\_ratio)} & Evaluates a Positive-Unlabeled (PU) learning model by fitting it to the training data, making predictions, and printing evaluation metrics. \\
                
                \texttt{main()} & Main function that orchestrates the loading of the dataset, preprocessing, model training, and evaluation. Allows toggling of evaluations for base ML models and PU learning models. \\ 
                \hline
            \end{tabular}
            \vspace{0.5cm}
            \caption{Overview of our Code's Functions and Classes}
            \label{tab:code_overview}
        \end{table*}
    
    \subsection{Machine Learning Models}
    
        We evaluate four established machine learning algorithms within our PU Learning framework:
        \begin{enumerate}
            \item Na\"{i}ve Bayes (NB) (\texttt{GaussianNB})
            \item Support Vector Machine (SVM) (\texttt{LinearSVC})
            \item Random Forest (RF) (\texttt{RandomForestClassifier})
            \item XGBoost (\texttt{XGBClassifier})
        \end{enumerate}
    
\subsection{Evaluation Metrics}
    
        To evaluate the performance of PU Learning in DDoS detection, we employ \textit{Bekker et al.'s} \cite{Bekker2020} modified \(F_1\) score, adapted for PU Learning scenarios:

        \[\text{PU } F_{1} \text{ approximation} \approx \frac{2pr}{p + r} \approx \frac{r^2}{Pr(\hat{\mathbf{y}} = 1)}\]

        Where:
        \begin{itemize}
            \item \(p = Pr(\mathbf{y} = 1 \mid \hat{\mathbf{y}} = 1)\): The conditional probability that an example is truly positive given that it is classified as positive.
            \item \(r = Pr(\hat{\mathbf{y}} = 1 \mid \mathbf{y} = 1)\): The conditional probability that an example is classified as positive given that it is truly positive.
            \item \(Pr(\hat{\mathbf{y}} = 1)\): The probability that an example is classified as positive.
        \end{itemize}

        To clearly illustrate what \(\hat{\mathbf{y}}\) is, we provide an example. Take a training set with the following variables: \texttt{Age}, \texttt{Familial Diabetes Diagnosis}, \texttt{Presence of Fatigue}, \texttt{Frequency of Urination Per Day}, and \texttt{the Presence of Blurred Vision}. From what little positive data samples a research team possesses, they determined that the following vector, $(63, \text{No}, \text{Yes}, 10, \text{No})$, corresponds to or is indicative of the presence of some arbitrary disease. As such, \(y\), or the indicator variable for an example to be positive, is set to 1. A value of 1 means that the thing we are trying to detect is present based on the previously defined vector, which is shorthanded to \(x\). As a result, \(\hat{\mathbf{y}}\) simply represents the collection of positive indicator variables, \(y\), with a value of 1. This set of indicator variables is tied to the vector of attributes \(x\). In this way, \(\hat{\mathbf{y}}\) captures the set of attributes, that when appraised in concert, leads to a positive conclusion, be that the existence of something or some other analogous example.

        This metric allows us to assess the performance of our PU Learning models in a context where traditional metrics like Precision and Recall may not be directly applicable due to the absence of labeled negative examples.

        \subsubsection*{Metrics for Evaluating Machine Learning Models in PU Learning}
        
            In the context of evaluating machine learning models, particularly within a Positive-Unlabeled (PU) learning approach, several key metrics provide insights into the performance of the computational techniques. Here, we discuss four important metrics: \(F_{1}\) Score, ROC AUC Score, Recall, and Precision.
            
            \begin{itemize}
                \item \textit{\(F_{1}\) Score}: Continuing from our earlier discussion, the \(F_{1}\) Score is the harmonic mean of Precision and Recall, providing a single metric that balances both concerns. It is especially useful when dealing with imbalanced datasets, such as in PU Learning, where positive instances may be scarce compared to unlabeled instances. The \(F_{1}\) Score takes into account both false positives and false negatives, making it a robust measure for evaluating model performance.
            
                \item \textit{Precision}: Precision measures the proportion of true positive predictions among all positive predictions made by the model. In a PU Learning context, where only positive and unlabeled instances are available, Precision indicates how many of the instances classified as positive (or belonging to the target class) are indeed correct. High Precision suggests that the model is good at avoiding false positives.
            
                \item \textit{Recall}: Recall, also known as sensitivity, measures the proportion of true positives correctly identified by the model out of all actual positive instances. In a PU setting, Recall assesses how well the model captures the actual positives when it has to infer from unlabeled data. A high Recall indicates that most of the relevant instances are being identified.
            
                \item \textit{ROC AUC Score}: The ROC AUC (Receiver Operating Characteristic - Area Under the Curve) score evaluates the model's ability to distinguish between the positive and negative classes across various thresholds. The AUC represents the likelihood that the model will rank a randomly chosen positive instance higher than a randomly chosen negative instance. In PU Learning, where negative instances may not be explicitly labeled, a high ROC AUC score indicates a model's effectiveness in distinguishing true positives from unlabeled instances.
            
            \end{itemize}
            
            These metrics collectively provide a comprehensive evaluation of model performance, particularly in scenarios characterized by the challenges of Positive-Unlabeled learning, where traditional metrics may not suffice.

\section{Results and Discussion}

    In this section, we will present and analyze the results of our experiments, comparing the performance of different PU Learning implementations using Na\"{i}ve Bayes (NB), Support Vector Machine (SVM), Random Forest (RF), and XGBoost algorithms. We will discuss the implications of these results for DDoS detection in cloud environments.

    \subsection{Performance Comparison}

        \begin{figure}[H]
            \centering
            \includegraphics[width=1\linewidth]{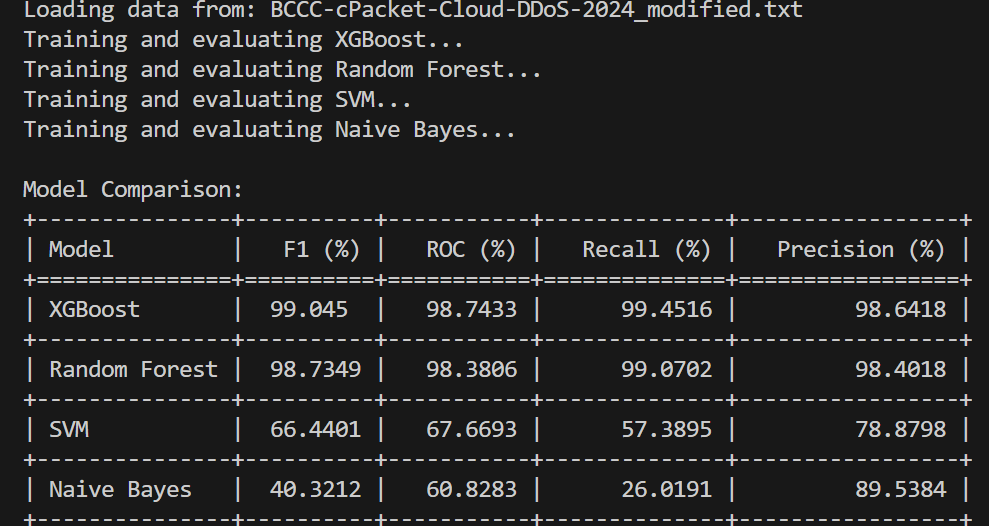}
            \caption{PU Learning Trial 1}
            \label{fig:trial_01}
        \end{figure}

        \begin{figure}[H]
            \centering
            \includegraphics[width=1\linewidth]{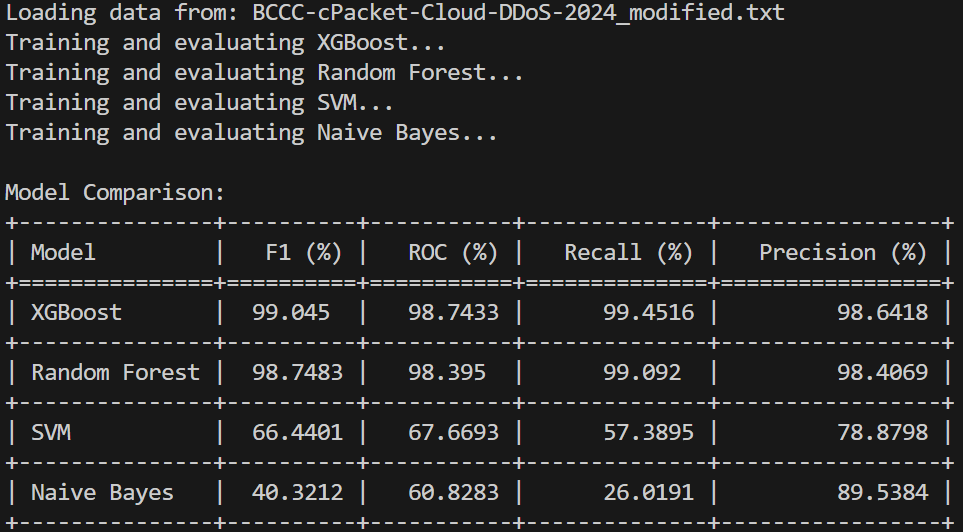}
            \caption{PU Learning Trial 2}
            \label{fig:trial_02}
        \end{figure}

        \begin{figure}[H]
            \centering
            \includegraphics[width=1\linewidth]{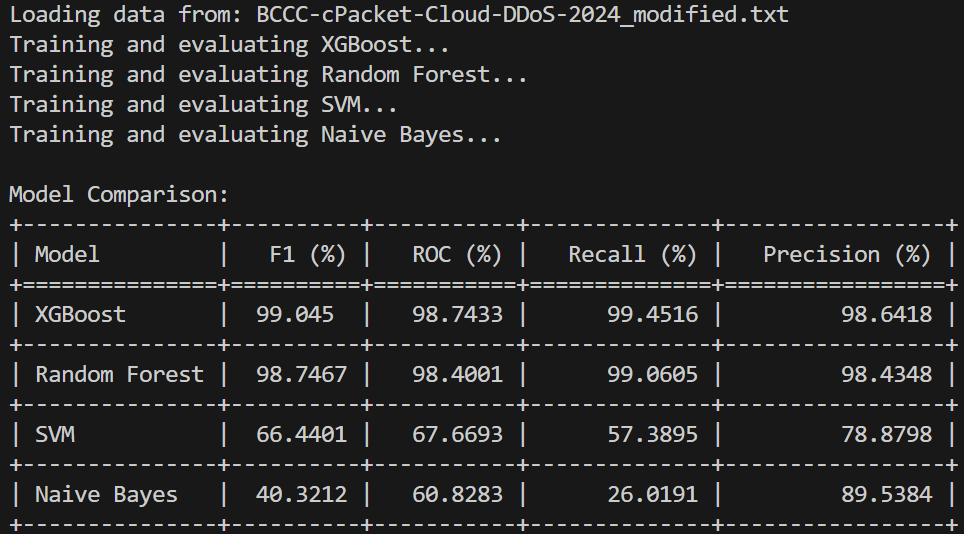}
            \caption{PU Learning Trial 3}
            \label{fig:trial_03}
        \end{figure}

        \begin{figure}[h!]
            \centering
            \includegraphics[width=1\linewidth]{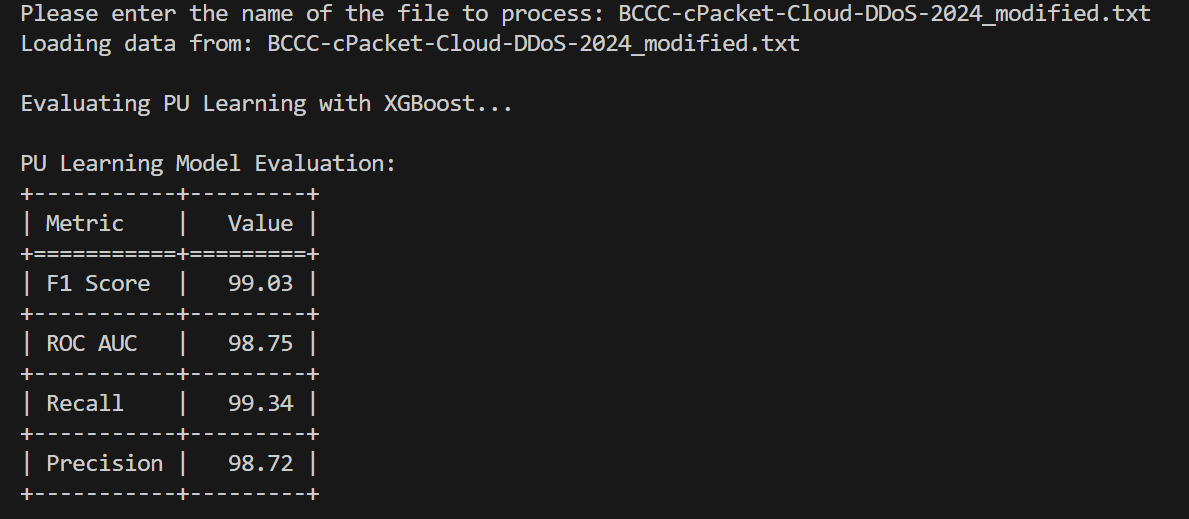}
            \caption{PU Learning Trial 4}
            \label{fig:trial_04}
        \end{figure}

        \begin{figure}[H]
            \centering
            \includegraphics[width=1\linewidth]{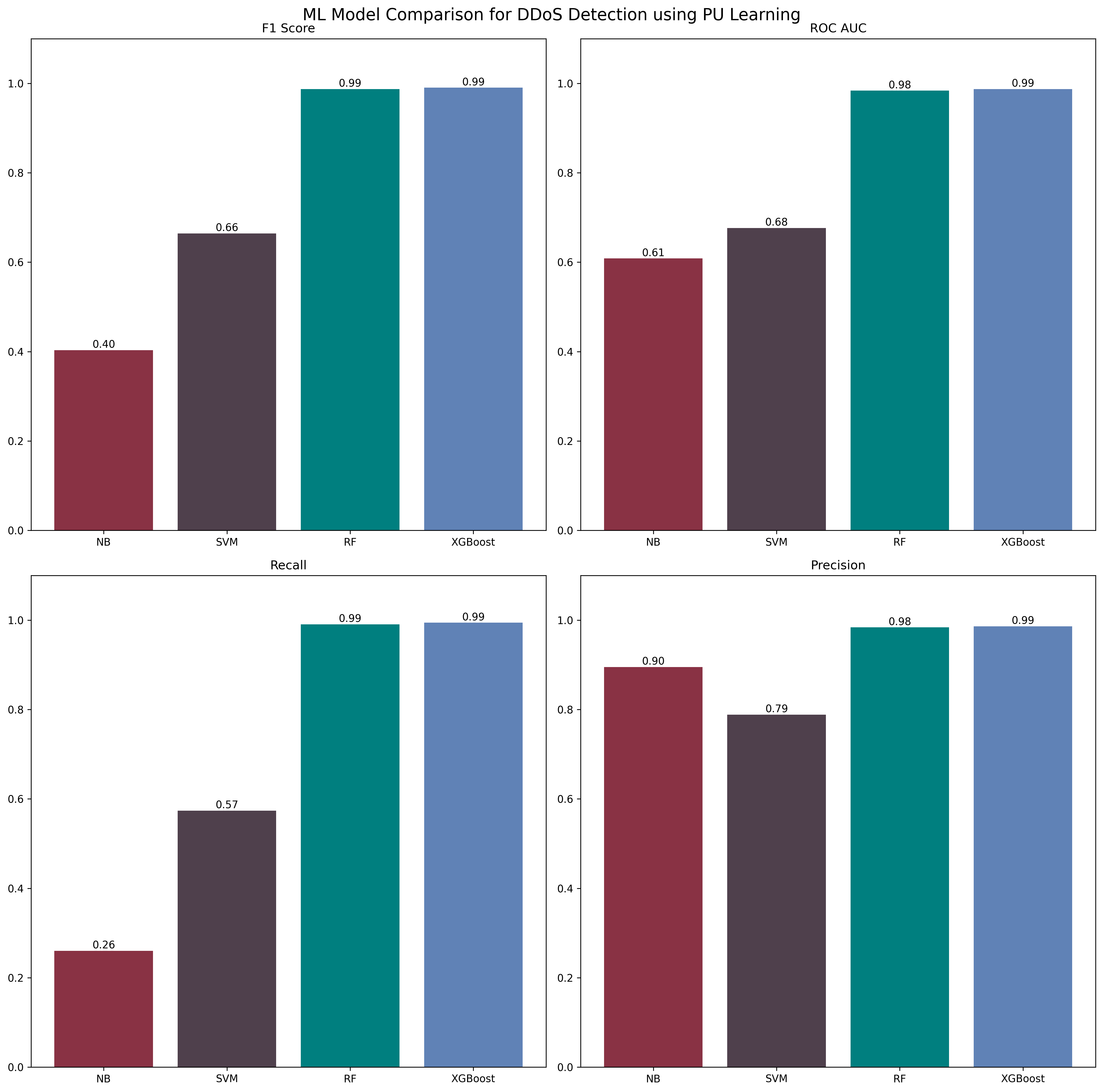}
            \caption{Comparison of Machine Learning (ML) Model Performance Metrics for DDoS Detection using PU Learning}
            \label{fig:ml_comparison}
        \end{figure}
        
    \subsection{Analysis of Results}
    
        The results of our PU Learning trials reveal significant variations in performance across the four machine learning algorithms tested: XGBoost, Random Forest, Support Vector Machine (SVM), and Na\"{i}ve Bayes.
        
        XGBoost demonstrated the best overall performance, achieving the highest scores across all metrics. Its \(F_{1}\) Score of 99.045, ROC AUC of 98.7433, Recall of 99.4516, and Precision of 98.6418 indicate exceptional ability to detect DDoS attacks while minimizing false positives and negatives.
        
        Random Forest closely followed XGBoost, with only slightly lower scores across all metrics (\(F_{1}\): 98.7349, ROC AUC: 98.3806, Recall: 99.0702, Precision: 98.4018). This suggests that ensemble methods are particularly effective for DDoS detection in PU Learning scenarios.
        
        SVM showed moderate performance, with an \(F_{1}\) Score of 66.4401 and ROC AUC of 67.6693. Its lower Recall (57.3895) but higher Precision (78.8798) suggest it may be more conservative in its predictions, potentially missing some attacks but having higher confidence in the ones it does identify.
        
        Na\"{i}ve Bayes showed mixed performance. It had the lowest \(F_{1}\) Score (40.3212) and Recall (26.0191), indicating it struggled to identify a large portion of the DDoS attacks. However, it demonstrated a high Precision (89.5384), suggesting that when it did classify an instance as a DDoS attack, it was correct nearly 90\% of the time. This high Precision but low Recall indicates that Na\"{i}ve Bayes might be overly conservative in its classifications, missing many attacks but having high confidence in the ones it does identify.
    
    \subsection{Implications for Cloud-based DDoS Detection}
    
        These results have several important implications for cloud-based DDoS detection:
        
        \begin{enumerate}
            \item Ensemble methods (XGBoost and Random Forest) appear to be the most effective for DDoS detection in PU Learning scenarios, likely due to their ability to capture complex patterns in network traffic.
            \item The high performance of XGBoost and Random Forest suggests that PU Learning can be highly effective for DDoS detection, even with limited labeled data.
            \item The poor performance of Na\"{i}ve Bayes indicates that simpler probabilistic models may not be suitable for the complex patterns involved in DDoS attacks.
            \item The moderate performance of SVM suggests that linear separation may not be sufficient for distinguishing between normal and DDoS traffic in all cases.
            \item The high Recall rates of XGBoost and Random Forest are particularly important for DDoS detection, as missing an attack (false negative) can have severe consequences for cloud services.
        \end{enumerate}

\section{Experimental Comparison with State-of-the-Art Approaches}

    In this section, we compare our PU Learning approach with a state-of-the-art DDoS detection model proposed by \textit{Shafi et al.} \cite{Shafi2024}. Their work introduces a sophisticated traffic characterization model that employs a multi-layered structure for distinguishing between normal and malicious network activities.
    
    \subsection{Shafi et al.'s Proposed Model}
    
        \textit{Shafi et al.} present a dual-path architecture that enhances flexibility and interpretability in DDoS detection. Their model consists of two primary layers:
        
        \begin{enumerate}
            \item An initial layer trained on labeled data to establish input classification.
            \item A second layer featuring two distinct models:
                \begin{itemize}
                    \item One trained exclusively with benign data
                    \item Another trained solely with attack data
                \end{itemize}
        \end{enumerate}
        
        The second layer's routing is determined by the first layer's classification: benign inputs are directed to the benign model, while attack inputs are sent to the attack model. This approach aims to mitigate overfitting and address the black-box nature often associated with deep learning-based algorithms.
    
    \subsection{Experimental Scenario Comparison}
    
        \textit{Shafi et al.} defined seven distinct experiment scenarios, with Task 6 being most comparable to our experiment. Task 6 focuses on identifying attack activities and the benign label, which aligns closely with our PU Learning approach for DDoS detection.
    
    \subsection{Methodology Comparison}
    
        Both our study and \textit{Shafi et al.'s} work employ four well-established machine learning algorithms: Na\"{i}ve Bayes (NB), Support Vector Machine (SVM), Random Forest (RF), and XGBoost. This commonality in algorithm selection allows for a meaningful comparison between the two approaches.
        
        The evaluation metrics used in both studies also show significant overlap, with both considering Precision, Recall, and \(F_{1}\)-Score. While our study additionally employs the ROC AUC metric, the shared metrics provide a solid basis for comparison.
    
    \subsection{Performance Comparison}
    
        To facilitate a clear comparison between\textit{ Shafi et al.'s} results and our PU Learning approach, we present their findings alongside ours in Table \ref{tab:performance_comparison}.
        
        \begin{table*}[h!]
            \centering
            \begin{tabular}{|l|c|c|c|c|c|c|}
            \hline
            \multirow{2}{*}{\textbf{Model}} & \multicolumn{3}{c|}{\textbf{\textit{Shafi et al.}}} & \multicolumn{3}{c|}{\textbf{Our PU Learning}} \\
            \cline{2-7}
             & \textbf{Precision} & \textbf{Recall} & \textbf{\(F_{1}\)-Score} & \textbf{Precision} & \textbf{Recall} & \textbf{\(F_{1}\)-Score} \\
            \hline
            NB & 0.58 & 0.29 & 0.19 & 0.8954 & 0.2602 & 0.4032 \\
            SVM & 0.35 & 0.50 & 0.40 & 0.7888 & 0.5739 & 0.6644 \\
            RF & 0.88 & 0.86 & 0.87 & 0.9840 & 0.9907 & 0.9873 \\
            XGBoost & 0.91 & 0.86 & 0.87 & \textbf{0.9864} & \textbf{0.9945} & \textbf{0.9905} \\
            Proposed & \textbf{0.97} & \textbf{0.96} & \textbf{0.97} & -- & -- & -- \\
            \hline
            \end{tabular}
            \vspace{0.5cm}
            \caption{Performance Comparison: \textit{Shafi et al.} vs. Our PU Learning Approach}
            \label{tab:performance_comparison}
        \end{table*}
        
        \begin{figure}[H]
        \centering
        \includegraphics[width=1\linewidth]{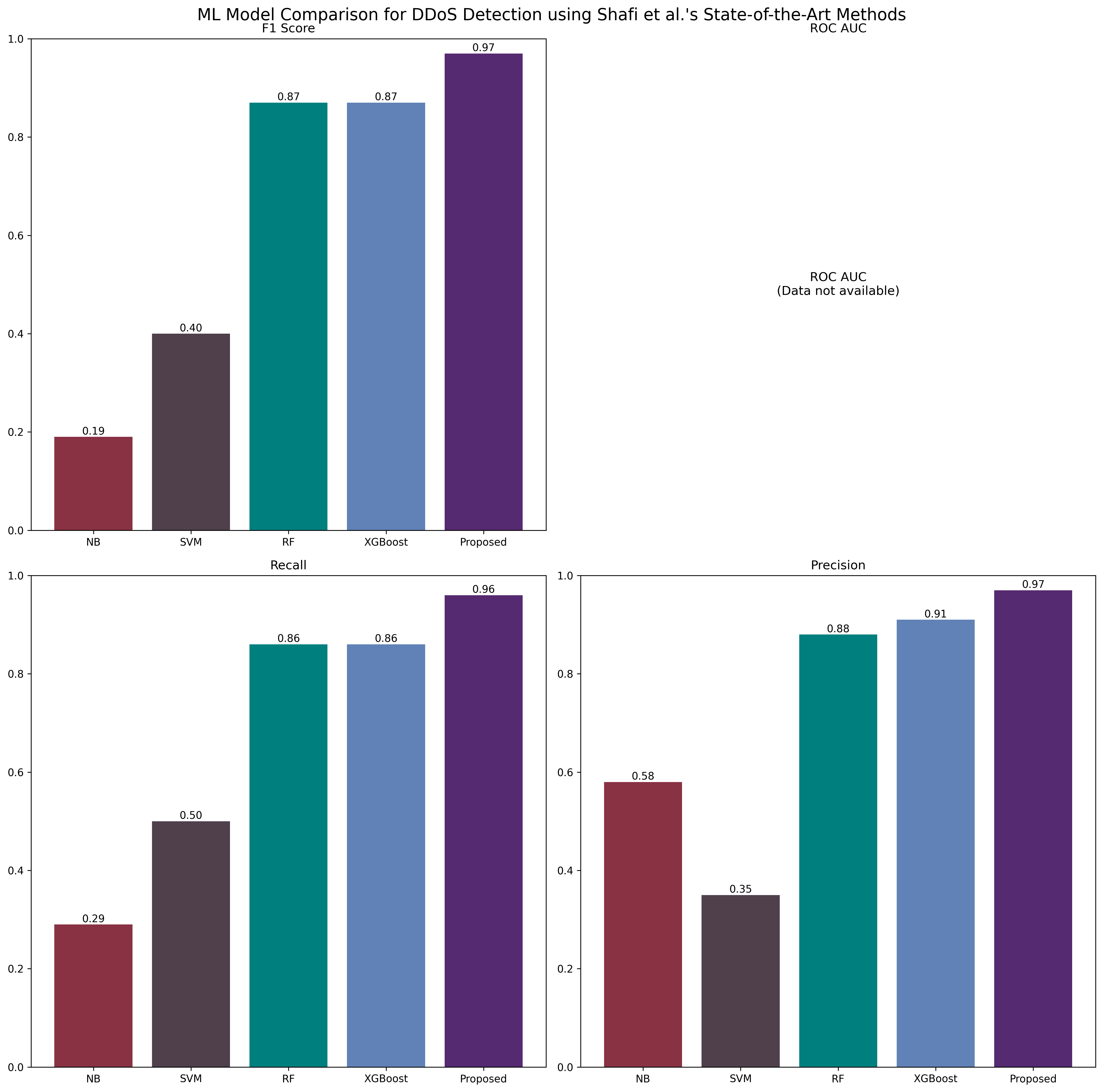}
        \caption{Performance Metrics for \textit{Shafi et al.'s} DDoS Detection Models}
        \label{fig:shafi_results}
        \end{figure}
    
        Analyzing the results, we observe several key points:
        
        \begin{enumerate}
            \item Both approaches demonstrate the superiority of ensemble methods (RF and XGBoost) over traditional algorithms (NB and SVM) for DDoS detection.
            \item Our PU Learning approach shows consistently higher performance across all metrics for each algorithm, particularly for NB and SVM.
            \item The performance gap between ensemble methods and traditional algorithms is more pronounced in our PU Learning approach.
            \item Both \textit{Shafi et al.'s} proposed model and our PU Learning implementation of XGBoost achieve the highest overall performance in their respective studies.
        \end{enumerate}
    
    \subsection{Comparative Analysis}
    
        While both approaches demonstrate high effectiveness in DDoS detection, our PU Learning method shows superior performance across all evaluated algorithms. This is particularly evident in the substantial improvements observed for NB and SVM, which struggle in \textit{Shafi et al.'s} implementation but perform notably better in our PU Learning framework.
        
        The exceptional performance of our PU Learning approach, especially with ensemble methods, suggests that it may be more adept at capturing the complex patterns inherent in DDoS attacks, even with limited labeled data. This capability is crucial in real-world scenarios where comprehensive labeled datasets are often unavailable or impractical to obtain.
        
        However, it's important to note that \textit{Shafi et al.'s} proposed model offers additional benefits, such as enhanced interpretability and a dual-path architecture that separates benign and attack data processing. These features could provide valuable insights in operational settings and may offer advantages in terms of model explainability.
    
    \subsection{Approach Determination}
    
        Based on the comparative analysis, our PU Learning approach demonstrates superior performance in terms of raw metrics. However, both approaches offer unique strengths. Our method excels in scenarios with limited labeled data and shows remarkable performance across various algorithms. \textit{Shafi et al.'s} model, while slightly lower in raw performance, offers enhanced interpretability and a specialized architecture for separating benign and attack traffic.
        
        The choice between these approaches may depend on specific use case requirements, such as the availability of labeled data, the need for model interpretability, or the importance of raw performance metrics. Future work could explore combining elements of both approaches to leverage their respective strengths in cloud-based DDoS detection systems.

\section{Contrastive Study of PU Learning and NU Learning Approaches in DDoS Detection}

    In this section, we conduct a comparative analysis between the PU Learning and NU Learning approaches for Cloud-based DDoS detection. By analyzing the differences in how each framework handles the labeling of data and the impact of these strategies on model performance, we aim to highlight the strengths and limitations of each approach.

    \subsection{Preprocessing and Binary Label Assignment}
    
        In our preprocessing stage, we assign a binary label to each data instance based on its prior classification as \texttt{Benign} or \texttt{!Benign}. If the prior data is labeled as benign, it receives a value of 0; otherwise, it receives a value of 1. This binary labeling system enables the classification of data into two distinct categories, which is foundational to the PU Learning framework that we apply in our problem setup.
    
    \subsection{PU Learning Problem Setup}
    
        Our problem setup is structured within the framework of PU Learning (Positive-Unlabeled Learning). In this context, instances labeled as \texttt{!Benign} represent the positive class (value of 1), while the benign instances are treated as the unlabeled class (value of 0). PU Learning is particularly suitable for scenarios in which only a subset of data instances is confidently labeled as positive, while the rest remain unlabelled rather than explicitly negative.
        
        Our PU Learning model implementation follows the two-step approach detailed earlier in this paper. First, the model trains on the positive samples (labeled as \texttt{!Benign}) and treats the unlabeled samples as potential positive examples. In the second step, it refines the model to better distinguish between true positives and false positives among the unlabeled data. This approach is crucial in instances where data is asymmetrically labeled, as is the case with our dataset of interest: \texttt{BCCC-cPacket-Cloud-DDoS-2024}.
    
    \subsection{NU Learning: Concept and Setup}
    
        NU Learning (Negative-Unlabeled Learning) is the converse of the PU Learning problem setup. In NU Learning, the positively labeled examples are treated as \say{unlabeled,} while the negative examples (originally \texttt{Benign}) are the primary focus for learning. In the context of our prior preprocessing, this would mean that instances labeled as 0 (benign) are now treated as the positive class, while those labeled as 1 (\texttt{!Benign}) are treated as the unlabeled class.
        
        The implementation of NU Learning involves a similar two-step approach, wherein the model trains on the available negative (benign) samples and progressively learns to distinguish between benign and potentially non-benign instances. NU Learning is applied to understand how well the model can classify an instance based on its alignment with known benign features.
    
    \subsection{Extending the Comparative Analysis}
    
        To deepen our analysis, we extend the study to compare the efficacy of framing Cloud-based DDoS detection as either a PU Learning or NU Learning problem. By examining both setups, we can assess which approach more accurately captures the nuanced patterns in the dataset and thus enables more effective detection of anomalous (or potentially malicious) instances.
    
    \subsection{Dataset Modification for NU Learning}
    
        To implement NU Learning, we modify the dataset by flipping the binary label values in the last column. Every instance of a 0 in the last column is changed to a 1 and vice versa. After this modification, we output the count of rows ending in 0 and 1 both before and after the transformation to confirm the accuracy of the modification. We then re-run our analysis code with the modified dataset to generate a new bar chart for the NU Learning approach, allowing for a direct comparison between the PU and NU Learning outcomes.
    
        \begin{figure}[H]
            \centering
            \includegraphics[width=1\linewidth]{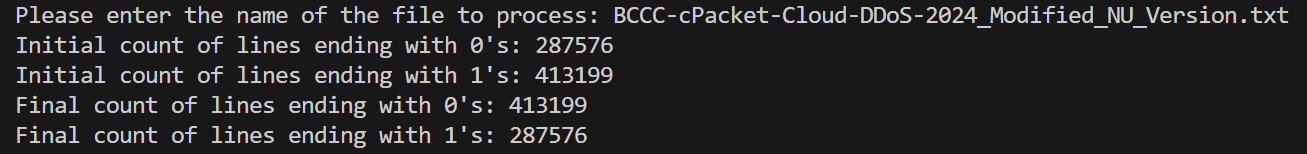}
            \caption{NU Learning Problem Recontextualization and Post Processing}
            \label{fig:nu_post_processing}
        \end{figure}
    
    \subsection{NU Learning Results}
    
        \begin{figure}[h!]
            \centering
            \includegraphics[width=1\linewidth]{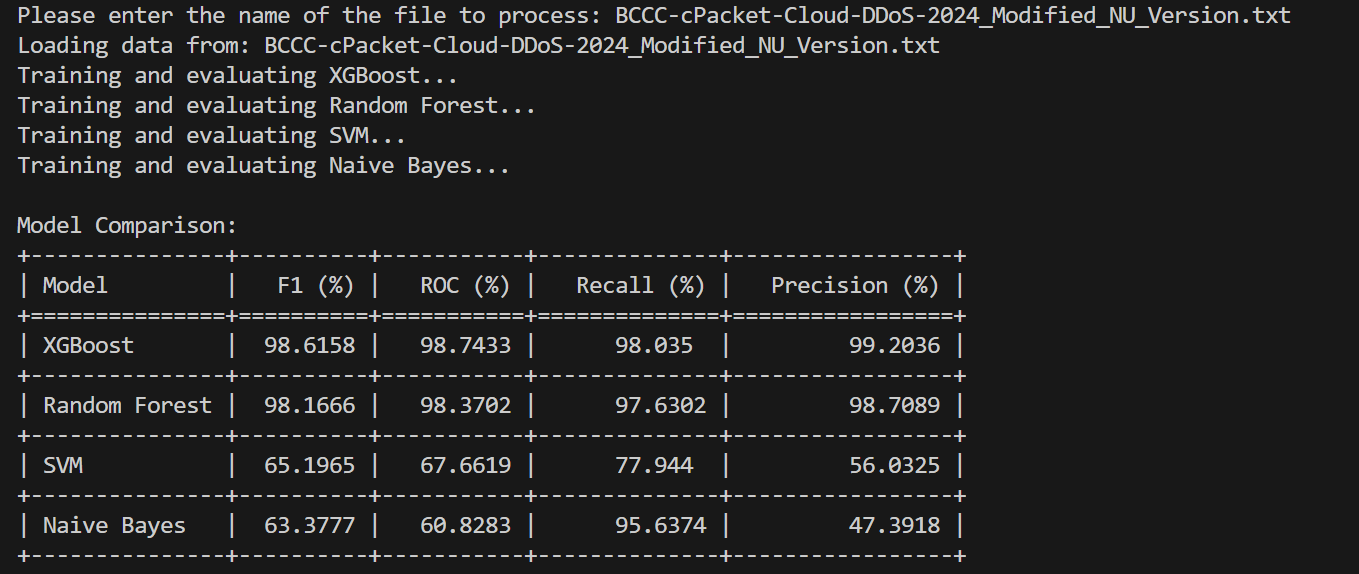}
            \caption{NU Learning Trial 1}
            \label{fig:nu_trial_01}
        \end{figure}

        \begin{figure}[h!]
            \centering
            \includegraphics[width=1\linewidth]{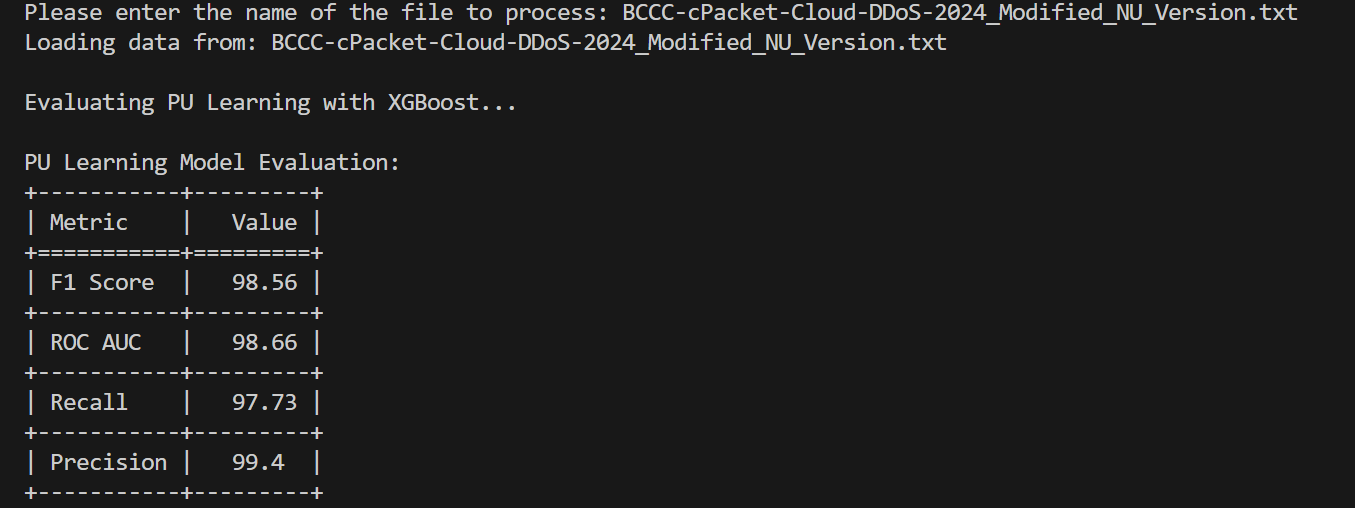}
            \caption{NU Learning Trial 2}
            \label{fig:nu_trial_02}
        \end{figure}

        \begin{figure}[H]
            \centering
            \includegraphics[width=1\linewidth]{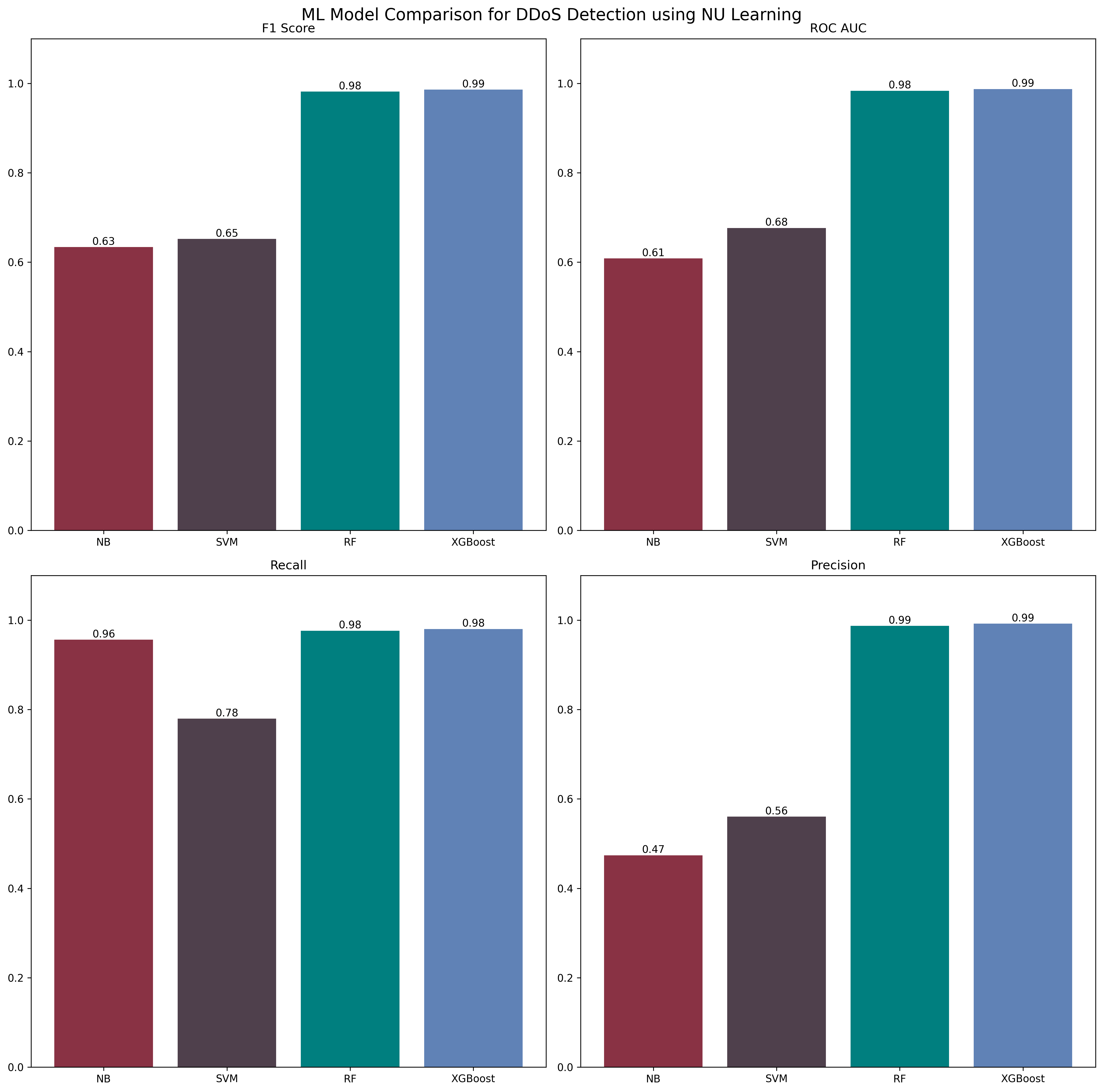}
            \caption{Comparison of Machine Learning (ML) Model Performance Metrics for DDoS Detection using NU Learning}
            \label{fig:ml_comparison_nu}
        \end{figure}

        \begin{table*}[h!]
            \centering
            \begin{tabular}{|l|c|c|c|c|}
                \hline
                \textbf{Model} & \textbf{\(F_{1}\)-Score} & \textbf{ROC AUC} & \textbf{Recall} & \textbf{Precision} \\ 
                \hline
                XGBoost & \textbf{0.986158} & \textbf{0.987433} & \textbf{0.980350} & \textbf{0.992036} \\
                Random Forest & 0.981666 & 0.983702 & 0.976302 & 0.987089 \\
                SVM & 0.651965 & 0.676619 & 0.779440 & 0.560325 \\
                Na\"{i}ve Bayes & 0.633777 & 0.608283 & 0.956374 & 0.473918 \\ 
                \hline
            \end{tabular}
            \vspace{0.5cm}
            \caption{Results of the NU-Learning Approach}
            \label{tab:nu_learning_results}
        \end{table*}

        \begin{figure}[H]
            \centering
            \includegraphics[width=1\linewidth]{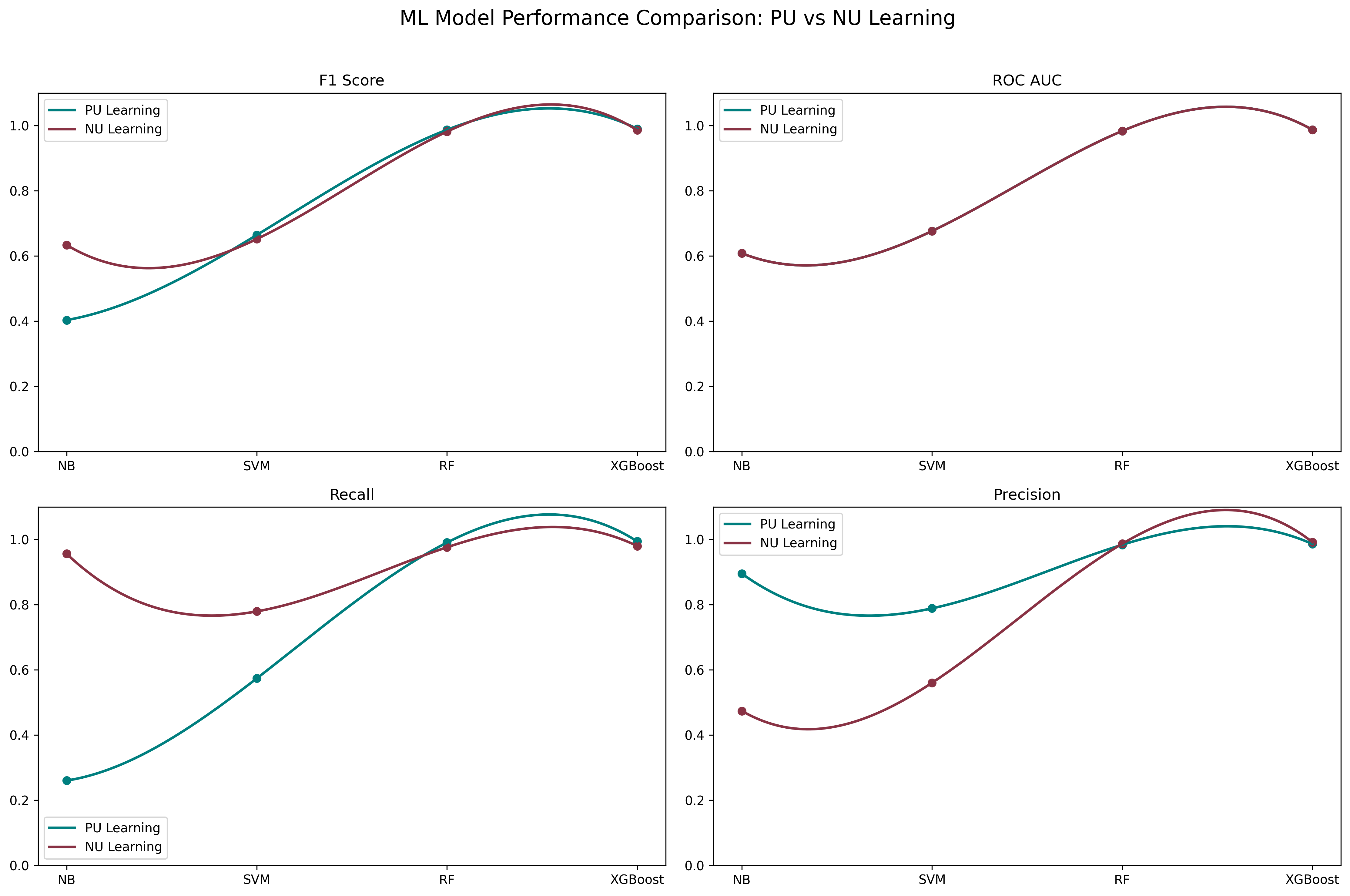}
            \caption{Evaluating Machine Learning Model Performance Metrics for DDoS Detection: PU Learning vs. NU Learning Problem Framing}
            \label{fig:ml_comparison_pu_vs_nu}
        \end{figure}

        In terms of the \textit{\(F_{1}\)-Score} (as exhibited in Figure \ref{fig:ml_comparison_pu_vs_nu}), both PU and NU Learning show consistent performance, with the exception of NU Learning's Na\"{i}ve Bayes implementation, which achieves a slightly higher score. When considering \textit{Recall}, NU Learning surpasses PU Learning, particularly with its Na\"{i}ve Bayes and Support Vector Machine (SVM) models. The performance of Random Forest and XGBoost, however, remains nearly identical across both learning frameworks. Similarly, the \textit{ROC AUC} scores for both models are virtually the same. On the other hand, PU Learning excels in \textit{Precision} when paired with Na\"{i}ve Bayes and SVM, while Random Forest and XGBoost once again demonstrate comparable results in both setups.
    
    \subsection{Assessment Deliberation}
    
        The comparison of PU and NU Learning results shows that both approaches yield high performance with the XGBoost and Random Forest algorithms. However, XGBoost consistently achieves higher precision and recall scores across both setups, indicating it is particularly effective for identifying the nuanced distinctions in DDoS attack patterns. Meanwhile, SVM and Na\"{i}ve Bayes underperform in both settings, suggesting that these algorithms may be less suitable for this problem due to their lower adaptability to the complex feature sets associated with DDoS detection.
        
        As an aside, it is infeasible for us to incorporate prior knowledge into the Na\"{i}ve Bayes implementation, which may account for its poorer performance. Similarly, the use of a linear support vector machine (SVM) likely contributes to SVM's limited efficacy in this context.
        
        Based on these findings, the PU Learning approach with the XGBoost algorithm is recommended for implementation, as it provides a more balanced and reliable model for distinguishing between benign and malicious traffic in Cloud-based DDoS detection tasks.

\section{Conclusion and Future Work}

    This study has explored the application of Positive-Unlabeled (PU) learning in the context of cloud-based DDoS detection, addressing a critical gap in the existing literature. By leveraging the \texttt{BCCC-cPacket- Cloud-DDoS-2024} dataset \cite{Kaggle2024} and implementing PU Learning with four established machine learning algorithms, we have demonstrated the potential of this approach in enhancing anomaly detection capabilities in cloud environments.

    Key findings of our study include:

    \begin{enumerate}
        \item Ensemble methods, particularly XGBoost and Random Forest, demonstrate superior performance in PU Learning-based DDoS detection, with \(F_{1}\) Scores exceeding 98\%.
        \item There is a significant performance gap between ensemble methods and traditional algorithms like SVM and Na\"{i}ve Bayes in this context.
        \item PU Learning shows great promise for effective DDoS detection in cloud environments, even with limited labeled data.
    \end{enumerate}

    These results highlight the promise of PU Learning in addressing the challenges of DDoS detection in cloud computing, particularly in scenarios where obtaining comprehensive labeled datasets is impractical or resource-intensive.

    \subsection{Contributions}

    This work has made several significant contributions to the field:

    \begin{enumerate}
        \item We have bridged the gap between PU Learning and cloud-based anomaly detection, demonstrating the applicability and effectiveness of this approach in DDoS detection.
        \item We have quantified the predictive and detective efficacy of PU Learning implementations using common machine learning methods, providing a comprehensive comparison of their performance.
        \item We have laid the groundwork for addressing Context-Aware DDoS Detection--which takes into account various contextual factors, such as user behavior, traffic patterns, application characteristics, and environmental conditions--in multi-cloud environments, a critical challenge in modern cloud security.
    \end{enumerate}

    \subsection{Future Work}

    While this study has made significant strides in applying PU Learning to cloud-based DDoS detection, several avenues for future research remain:

    \begin{enumerate}
        \item Exploring the application of PU Learning to other types of cloud security threats beyond DDoS attacks.
        \item Conducting a more in-depth analysis of feature importance in the high-performing models to gain insights into the most relevant indicators of DDoS attacks in cloud environments.
        \item Developing adaptive PU Learning models that can evolve with changing attack patterns and cloud infrastructure configurations.
        \item Extending the research to real-time DDoS detection scenarios, considering the computational efficiency and scalability of PU Learning approaches in production environments.
    \end{enumerate}

    In conclusion, this work has demonstrated the potential of PU Learning in enhancing cloud-based DDoS detection capabilities. As cloud computing continues to evolve and face increasingly sophisticated security threats, approaches like PU Learning will play a crucial role in developing more robust and adaptive defense mechanisms.

\bibliographystyle{ACM-Reference-Format}
\bibliography{references}

\end{document}